\documentclass[a4paper,conference]{IEEEtran}
\usepackage{cite}

\ifCLASSINFOpdf
\else
\fi
\usepackage{amsmath}
\usepackage{algorithmic}

\usepackage{array}

\usepackage{stfloats}
\usepackage{amssymb}
\usepackage{url}            
\usepackage{booktabs}       
\usepackage{amsfonts}       
\usepackage{nicefrac}       
\usepackage{microtype}      
\usepackage{graphicx}
\usepackage[pagebackref=true,breaklinks=true,colorlinks,bookmarks=false]{hyperref}

\usepackage{url}

\begin{document}
%
\title{$P^2$ Net: Augmented Parallel-Pyramid Net for Attention Guided Pose Estimation}

\author{\IEEEauthorblockN{Luanxuan Hou$^{1,2}$, Jie Cao$^{1,2}$, Yuan Zhao$^{3}$, Haifeng Shen$^{3}$, Jian Tang$^{3}$, Ran He\IEEEauthorrefmark{4}$^{1,2}$ }
\IEEEauthorblockA{$^{1}$University of Chinese Academy of Sciences\quad{$^2$}NLPR$\;\&\;$CEBSIT, CASIA\quad{$^3$}AI Labs, Didi Chuxing\\
{\small \{luanxuan.hou, jie.cao\}@cripac.ia.ac.cn, \{zhaoyuanjason, shenhaifeng, tangjian\}@didiglobal.com, rhe@nlpr.ia.ac.cn}}
}

%


\maketitle
\renewcommand{\thefootnote}{}
\footnote{\IEEEauthorrefmark{4} Corresponding author.}
\begin{abstract}
The target of human pose estimation is to determine the body parts and joint locations of persons in the image. Angular changes, motion blur and occlusion in the natural scenes make this task challenging, while some joints are more difficult to be detected than others. In this paper, we propose an augmented Parallel-Pyramid Net ($P^2Net$) with feature refinement by dilated bottleneck and attention module. During data preprocessing, we proposed a differentiable auto data augmentation ($DA^2$) method. We formulate the problem of searching data augmentaion policy in a differentiable form, so that the optimal policy setting can be easily updated by back propagation during training. $DA^2$ improves the training efficiency. A parallel-pyramid structure is followed to compensate the information loss introduced by the network. We innovate two fusion structures, i.e.  Parallel Fusion and Progressive Fusion, to  process  pyramid features from backbone network. Both fusion structures leverage the advantages of spatial information affluence at high resolution and semantic comprehension at low resolution effectively. We propose a refinement stage for the pyramid features to further boost the accuracy of our network. By introducing dilated bottleneck and attention module, we increase the receptive  field  for  the  features  with  limited  complexity and tune the importance to different feature channels. To further refine the feature maps after completion of feature extraction stage, an Attention Module ($AM$) is defined to extract weighted features from different scale feature maps generated by the parallel-pyramid structure. Compared with the traditional up-sampling refining, $AM$ can better capture the relationship between channels. Experiments corroborate the effectiveness of our proposed method. Notably, our method achieves the best performance on the challenging \textit{MSCOCO} and \textit{MPII} datasets.
\end{abstract}

\IEEEpeerreviewmaketitle

\section{Introduction}
Multi-person pose estimation has been intensely investigated in computer vision. It is a challenging task due to occluded, self-occluded, and invisible keypoints. There have been growing interests in developing deep neural network methods for multi-person pose estimation. In general, these methods can be categorized into top-down and bottom-up methods.

The top-down methods firstly detect each person in the image by applying detection algorithms, e.g., Feature Pyramid Network \cite{szegedy2017inception}, Mask RCNN \cite{he2017mask}, TridentNet \cite{li2019scale}, and then generate keypoints in these detected bounding-boxes. The representative methods include Simple-Baseline \cite{xiao2018simple}, Regional multi-person pose estimation \cite{fang2017rmpe}, Cascaded Pyramid Network \cite{chen2018cascaded}, and High-Resolution Network \cite{sun2019deep}. On the other hand, the bottom-up methods consist of keypoints detection and clustering. They directly predict all keypoints and then assemble them into the poses of each person. Part Affinity Field \cite{cao2017realtime}, Associative Embedding \cite{newell2017associative}, Part Segmentation \cite{xia2017joint}, Mid-Range offsets \cite{papandreou2018personlab} fall into this part.

Although great progress has been made, there are still many ongoing issues for pose estimation in the wild. In the unconstrained conditions, the visibility of keypoints is greatly affected by wearing, posture, viewing angle, background etc. The large pose variations further increase the difficulty in detection. Therefore, data augmentation plays an important role in pose estimation. Through data augmentation, we can diversify the data distributions to provide more samples in various scenarios. Existing augmentation strategies are usually hand-crafted or based on tuning techniques \cite{he2016deep,hu2018squeeze}. Due to the complexity of multi-person pose estimation, manually designed data augmentation operators is non-trivial. Moreover, unreasonably setting the data augmentation parameters will instead cause noisy interference. Hence, finding an efficient strategy becomes urgent. The existing methods of automatic search data augmentation sequences are all based on reinforcement learning (RL). A controller RNN predicts an augmentation policy from the search space. A sub-network (e.g., object detection and pose estimation) with a fixed architecture is trained to converge. Sub-network uses the accuracy during training as reward in reinforcement learning (RL). The reward $R$ will be used with the policy gradient method to update the controller so that it can generate better policies over time. Reinforcement learning (RL) is exploited to optimize the controller parameters. Though promising results are achieved, AutoAugment based on RL is extremely costly due to the low efficiency of RL and multi-pass training.

In this paper, we propose a novel automated approach, differentiable auto data augmentation ($DA^2$), to search data augmentation strategies which are explicitly suited to the pose estimation task. We design a new pose search space to encode the sequences of data augmentation that is commonly used in pose estimation. Particularly, our $DA^2$ method equips with the differentiable method instead of RL, making the search results particularly suitable for pose estimation. The data efficiency of gradient-based optimization, as opposed to inefficient black-box search, allows our approach to achieve competitive, if not better performance compared with the state of the art with orders of magnitude less computation resources. Because our auto data augmentation is associated with the actual target task, it also reduces the data noise caused by setting unreasonable data augmentation parameters. Because as far as we know, if we adopt a reasonable data augmentation strategy, we can balance the data distribution. And we can balance the distribution of noisy data and reduce the long-tail effect of the data distribution. These data augmentation methods can improve the fault tolerance of Convolutional Neural Networks (CNNs) for noisy data. However, if the inappropriate parameter adjustment is set, it will bring noisy data to the training of the model, which will negatively affect the performance of the model and reduces performance. On the other hand, automatic data augmentation strategies improve robustness enabling the trained model to work under various task. Moreover, the adaptability to unconstrained environments and the training robustness of noisy sample data are improved.

For the backbone network, we propose a novel two-stage network structure based on the top-down pipeline. First, we adopt parallel structures to learn high-resolution representations to compensate the information loss which is caused by the network downsample and feature pyramid network. On one hand, our parallel structures extract abundant context information necessary for the inference of the occluded and invisible keipoints. On the other hand, it effectively preserves the spatial and semantic information of the pyramid feature structure . Second, based on the pyramid features, our refinement stage explicitly addresses the hard keypoint detection problem by optimizing an \textit{Online-Hard-Keypoints-Mining Loss} \cite{shrivastava2016training}. The existing refining operation is to perform upsampling and then concatenation, which ignores the relationships between feature maps with different scales. Alternatively, we use the dilated bottleneck structure to increase the receptive field without increasing the amount of calculation. Meanwhile, in order to make refinement stage focus on more informative regions, we add the dilated bottleneck and an Attention Module ($AM$) to the output of $Feature~Extraction~Stage$. Our approach not only maintains high-resolution feature maps but also keeps large receptive field, both of which are important for pose estimation.

We summarize our contributions as follows:
\begin{itemize}
\item To the best of our knowledge, we are the first work doing adaptive data augmentation for human pose estimation problem. The data augmentaion policy is learnt to suit specific tasks and training data. Therefore the learnt model is more robust in testing. We formulate the problem of searching data augmentaion policy in a differentiable form, so that the optimal policy setting can be easily updated by back propagation during training. 
\item We innovate two fusion structures, i.e. Parallel Fusion and Progressive Fusion, to process pyramid features from backbone network. Both fusion structures leverage the advantages of spatial information affluence at high resolution and semantic comprehension at low resolution effectively.  
\item We propose a refinement stage for the pyramid features to further boost the accuracy of our network. By introducing dilated bottleneck and attention module, we increase the receptive field for the features with limited complexity and tune the importance to different feature channels. Our ablation study shows this refinement stage has a obvious contribution to the overall performance.
\item We conducted extensive experiments on the mainstream datasets and achieved the state-of-the-art single model performance. Remarkably, we got 77.3 AP results in the MSCOCO test-dev dataset with smaller Params and GFLOPs.

\end{itemize}
\section{Related Works}
In general, there are two mainstream approaches: Top-Down and Bottom-Up.

{\bf Top-Down methods:} Top-Down methods simulate the process of the human perception system. They first apply human detection, and then generate keypoints within the bounding boxes of human bodies. Some well-known human detectors are FPN \cite{lin2017feature}, YOLO \cite{redmon2016you}, SSD \cite{liu2016ssd} etc. After detection, in general, U-shape stacked hourglass network \cite{newell2016stacked} is used to predict the keypoints. However, when the person in the image is occluded or partially occluded, the prediction of some keypoints may fail. Therefore, applying an additional refinement step after the prediction may be beneficial.   

{\bf Bottom-Up methods:} Bottom-Up methods firstly predict the body joints and then group them into set of poses. Without replying on the output human detectors, bottom-up methods are generally much faster than top-down methods. Therefore they are broadly used in real-time scenarios. 

 Pishchulin \cite{insafutdinov2016deepercut} proposed a bottom-up approach. First, they used CNN to extract body part candidates. Each candidate area corresponded to a joint point, and each joint point was used as a node in the graph. The joint points (nodes in the graph) belonging to the same person were classified into one category, and each person was treated as a separate category. At the same time, another branch marked the detected nodes which part of the human body they belonged to. Finally, they used the classified persons combined with the marked parts to estimate the final pose  of each person. The method of Zhe \cite{cao2017realtime} predicted the heat map and Part Affinity Fields(PAFs) of keypoints of the input image. They then performed bipartite matching based on keypoints and body parts, and finally got all the poses of everyone in the picture. Newell \cite{newell2017associative} simultaneously produced score maps and pixel-wise embedding to group the candidate keypoints to different people to get final multi-person pose estimation. 


 However, bottom-up methods are usually less accurate in predicting body joints without the constraints of bounding boxes and may divide the body parts of the same person into different clusters. 

{\bf Auto Data Augmentation:} 
Data augmentation is very important to train robust networks for many tasks, e.g. object detection, pose estimation etc. Moreover, tailoring the augmentation policy to a specific task is usually more effective than applying a general one. However, manually selecting the augmentation policy is hard. First, the policy space is usually very large and unrealistic to manually select the best one. Second, if unfortunately a bad augmentation policy is selected, the trained network may perform the same or even worse comparing with network without data augmentation. Therefore, some researchers proposed  automatic searching methods to find a helpful data augmentation policy. A policy includes many sub-policies, each of which includes two parameters: probability and magnitude of the operation, e.g. rotation, translation etc. In AutoAugment\cite{cubuk2018autoaugment},  they use reinforcement learning (RL) to train a RNN controller  to predict the policy. Though promising results are achieved in \cite{cubuk2018autoaugment}, using RL for policy selection is extremely computationally demanding (e.g., 15000 GPU hours on ImageNet). In \cite{lin2019online,lim2019fast,ho2019population,he2009robust}, they improve the actual application effect and reduced the loss of network information.

\section{Proposed Method}

\begin{figure*}[htb]
    \centering
    \includegraphics[scale=0.48]{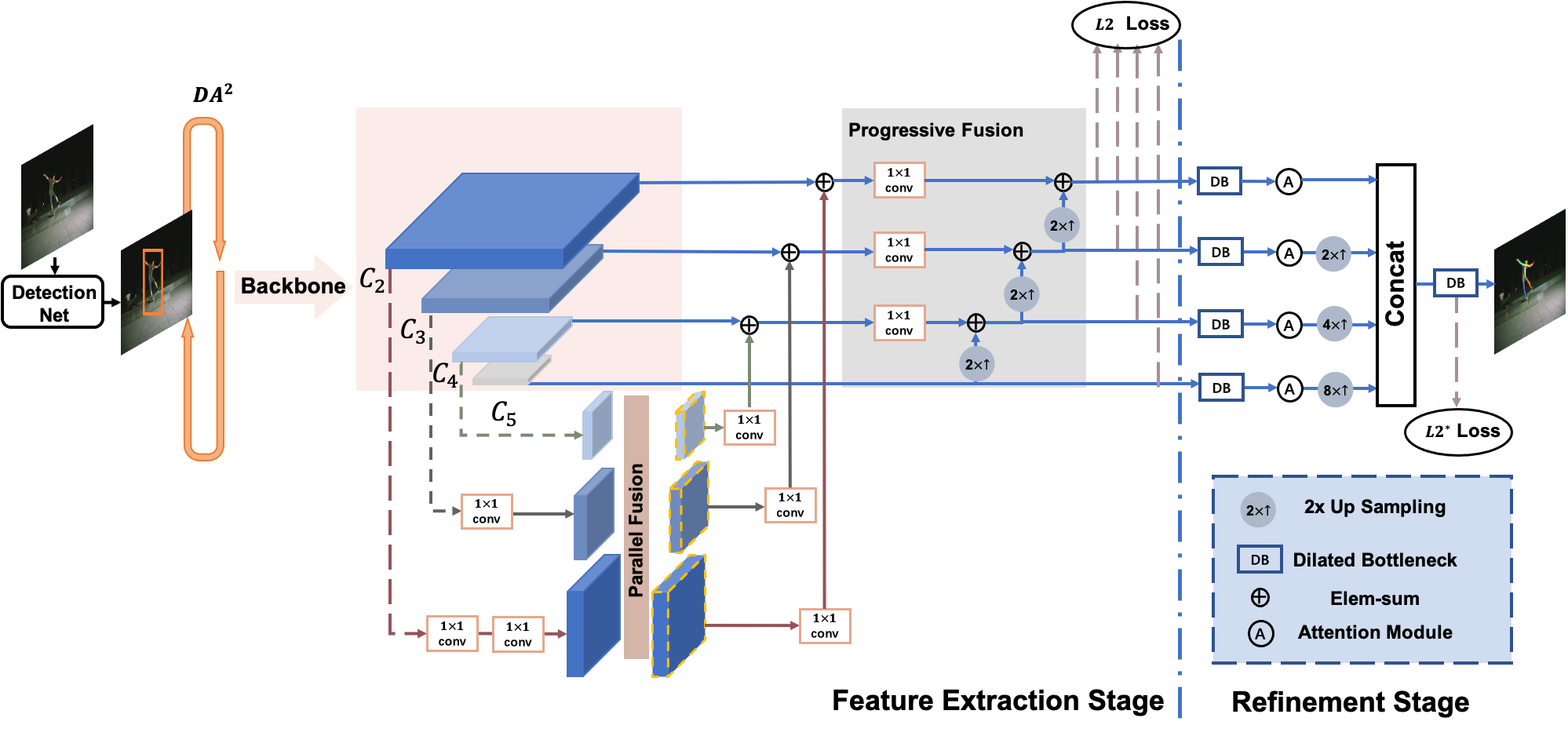}
    \caption{The architecture of the proposed $P^2Net$ network.}
    \label{fig.1}
\end{figure*}

Our proposed $P^2$ Net is a top-down method. We first apply a human detection network (i.e. DetNet) on the input image to obtain bounding boxes of the persons. Then we input the cropped images based on the bounding boxes to the $P^2$ Net to generate the heat maps of body parts. As shown in Figure~\ref{fig.1}, in training time, we introduce differentiable automatic data augmentation ($DA^2$) to generate training data from the the cropped images. In $P^2$ Net, it includes two stages: feature extraction stage and refinement stage. 

\subsection{Differentiable Automatic Data Augmentation}
\begin{figure}[!t]
    \centering
    \includegraphics[width=3.4in]{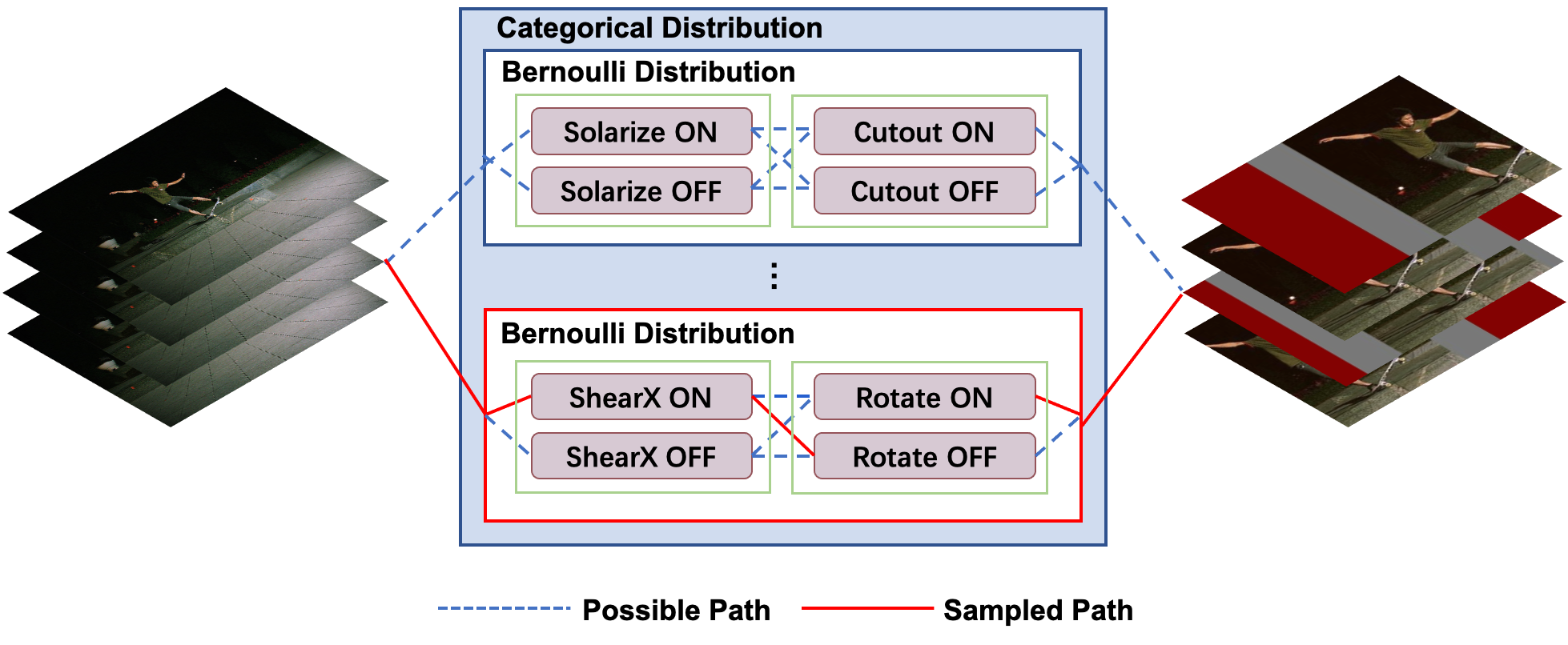}
    \caption{Example of sampled path and possible path in learning the the policy of automatic data augmentation.}
    \label{DADA}
\end{figure}

The search space of our data augmentation policy contains K sub-policies with each sub-policy consisting of N operations (e.g. translation, rotation etc.) applied in sequence to a single cropped image. Instead of treating each sub-policy equally, we aim to learn how likely one sub-policy should be used in the augmentation process. Inside the sub-policy, we further set two hyperparameters in each operation, probability \textit{p} of using the operation $O$ and magnitude \textit{m} (e.g. rotation angle) to control the operation sequence. In Figure~\ref{DADA}, the red line describes an example sampled path. When the k-th sub-policy $S_k$ is selected, the operation $ShearX$ is used when applying the probability 0.7 and magnitude 5, and followed by the second operation $Rotate$ bypassed (because the probability of $Rotate$ is 0.2). 

\subsubsection{Description of the sampled path}
Let $C$ be a Categorical variable with sub-policy probabilities $\pi_1$,$\pi_2$,...$\pi_K$. 
\begin{equation}
C = one\_hot(\arg_i max (\pi_i)), 
\end{equation}
When training, we want to sample a sub-policy from the fixed $Categorical~Distribution$ {$\pi_1$,$\pi_2$,...$\pi_K$} with some randomness. A simple and efficient way to do so is by applying the Gumbel-Max trick \cite{jang2016categorical}
\begin{equation}
C = one\_hot(\arg_i max (\log \pi_i + g_i)), 
\label{sample_C}
\end{equation}
where $g_i = -\log(-\log(u_i))$ are i.i.d samples drawn from $Gumbel(0,1)$. Nevertheless, the description of $C$ in Eq. \ref{sample_C} is discrete. We can't update $C$ using back propagation. Thus we use a generalized form of softmax as continuous and differentiable approximation to $arg max$.
\begin{equation}
    c_{k}=\frac{\exp((\log \pi_{k}+g_k)/\tau) }{\sum_{k'=0}^{K}\exp((\log \pi_{k'}+g_{k'})/\tau_1)}
\label{dist_pi}
\end{equation}
$\tau_1$ is a temperature parameter. The output of the augmentation is then formulated as:
\begin{equation}
\hat{x} = \sum_{k}c_{k}S_{k}(x)
\end{equation}


\subsubsection{Description of the sampled operation}
Each operation can be turned on or off inside a sub-policy, which resembles a Bernoulli variable, i.e. $b$ in Eq. \ref{bi_operation}. Here we ignore the index of operation $O$, operation probability $p$ and operation magnitude $m$ for simplicity of explanation. 
\begin{equation}
    \overline{O}(x,p,m)=bO(x,m)+(1-b)x.
\label{bi_operation}
\end{equation}
where $P(b=1)=p$. 

Similar to sampling the $Categorical~Distribution$, in training time, we want to generate random samples of $b$ from the $Bernoulli~Distribution$. After applying the Gumbel-Max trick \cite{maddison2016concrete}, we have the sample of $b$ as
\begin{equation}
    b = H(\log \frac{p}{1-p} + \log \frac{u}{1-u})
\label{step_b}
\end{equation}
where $u\sim Uniform(0,1)$ and $H$ is the unit step function. To make $b$ continuous and differentiable, we relax Eq. \ref{step_b} using a generalized form of $Sigmoid$ function with temperature parameter $\tau_2$
\begin{equation}
    b = \sigma((\log \frac{p}{1-p} + \log \frac{u}{1-u})/\tau_2).
\label{sigmoid_b}
\end{equation}
Calculating the gradient of $\overline{O}(x,p,m)$ w.r.t the relaxed formed of $b$ is straightforward. However, the gradient of $\overline{O}(x,p,m)$ w.r.t the other hyperparameter magnitude $m$ is not always straightforward, depending on the operation type. $m$ in some operations, like flipping, is not differentiable. Therefore we set the straight-through gradient estimator \cite{bengio2013estimating} for $m$ to be a constant, 
\begin{equation}
    \frac{\partial \overline{O}(x,p,m)}{\partial m}= const .
\end{equation}

\subsubsection{Optimization of data augmentaion}

To find the optimal set of sub-policies, we use bi-level optimization to learn both the parameter set $d={\pi, p, m}$ and the network weights $\omega$. Let $\mathcal{L}_{train}$ and $\mathcal{L}_{val}$ represent the training and validation loss respectively. Given $d$, we want to minimize $\mathcal{L}_{train}$ to obtain the optimal network, while given the network $\omega$, we want to minimize $\mathcal{L}_{val}$ to update the augmentaion setting $d$, i.e.:

\begin{equation}
\begin{aligned}
    \underset{d}{min}\mathcal{L}_{val}(\omega^\ast(d)) \\\omega^\ast(d)=argmin_\omega\mathcal{L}_{train}&(\omega,d).
\end{aligned}
\end{equation}

Note that, since we have modeled and relaxed the formulation of sampled path and sampled operation, $d$ can be updated through gradient descent, just like $\omega$. To speed up the optimization process, we alternate the learning of $\omega$ and $d$ at mini-batch level. At step k, given the data augmentation setting $d_{k-1}$, we train and update $\omega_k$ according to the training loss $E[\mathcal{L}_{train}(\omega_{k-1},d_{k-1})]$ over the mini-batch. Then we update $d_{k-1}$ to $d_k$ on a validation batch through $\mathcal{L}_{val}(\omega_k)$. We iterate the update of $d$ and $\omega$ until convergence.
At step k, when update $w_k$,
\begin{equation}
    \omega_{k} = \omega_{k-1}-\zeta\nabla_{\omega_{k-1}}\mathcal{L}_{train}(\omega_{k-1},d_{k-1}). 
\end{equation}
Applying the chain rule, the gradient of $\mathcal{L}_{val}(\omega_k)$ with respect to $d_{k-1} $ is 
\begin{equation}
    -\zeta\nabla_{\omega_{k-1},d_{k-1}}^2E[\mathcal{L}_{train}(\omega_{k-1},d_{k-1})]\nabla_{\omega_{k}}\mathcal{L}_{val}(\omega_{k}).
\label{ori_grad_d}
\end{equation}
The gradient in Eq. \ref{ori_grad_d} is computationally costly. Fortunately, we can save the complexity by using the finite difference approximation. Let $\epsilon$ be a very small scalar, $\omega_{k-1}^{+}=\omega_{k-1}+\epsilon \nabla_{w_{k}}\mathcal{L}_{val}(w_k)$ and $\omega_{k-1}^{-}=\omega_{k-1}-\epsilon\nabla_{w_{k}}\mathcal{L}_{val}(w_k)$. For simplicity, we ignore the index $k-1$ in $d_{k-1}$ in the following approximation form, i.e. the approximated gradient: 
\begin{equation}
     -\zeta\frac{\nabla_{d}\mathit{E} \left [ \mathcal{L}_{train}(\omega_{k-1}^{+},d)\right ]-\nabla_{d}\mathit{E}\left [\mathcal{L}_{train}(w_{k-1}^{-},d) \right ]}{2\epsilon }.
\end{equation}

In our experiment, we set $\epsilon =0.01/\begin{Vmatrix}
\nabla_{\omega_{k}}\mathcal{L}_{val}(\omega_{k})
\end{Vmatrix}_2$.


\subsection{Feature extraction with pyramid and parallel structures}
\textbf{Pyramid Features.} After we get the augmented input, the image will go through ResNet to get feature expressions from the end of conv2\_x ( $C_2$), conv3\_x ($C_3$), conv4\_x ($C_4$), conv5\_x ($C_5$) layers respectively. The resolution of features from the early layers to later layers is downscaled by 2 in each dimension layer by layer, e.g. resolution at $C_2$ is two times of the resolution at $C_3$ in both width and height. Therefore, the collection of those features has the pyramid characteristic. To leverage the advantages of both spatial information affluence and semantic comprehension, we further process the features from different layers by parallel fusion and progressive fusion at different resolutions. 

\begin{figure}[!t]
    \centering
    \includegraphics[width=3.4in]{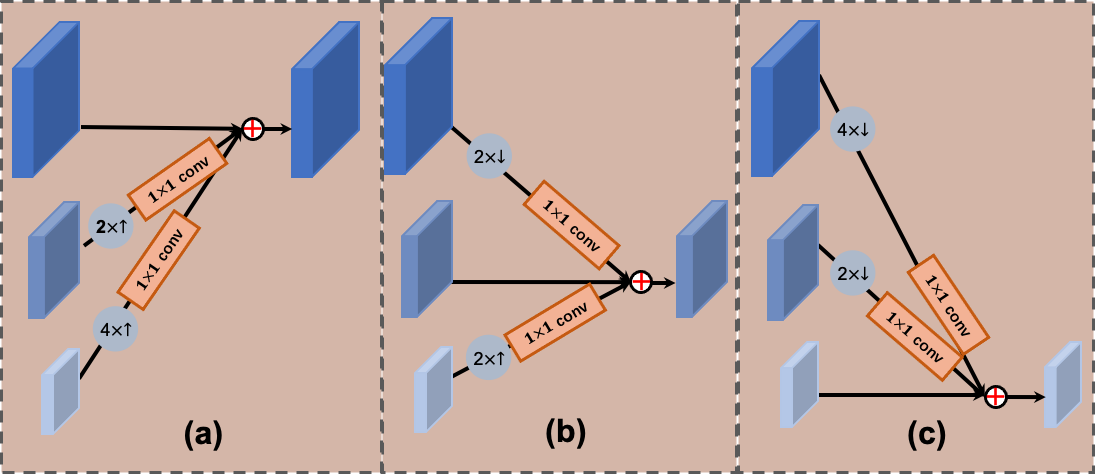}
    \caption{Parallel Fusion. (a) Fuse features at resolution of $C_2$. (b) Fuse features at resolution of $C_3$. (c) Fuse features at resolution $C_4$. }
    \label{fig.3}
\end{figure}
\textbf{Parallel Fusion.} As shown in Fig.~\ref{fig.1}, the dashed lines means data copy. We first apply a few more convolutions for $C_2$ and $C_3$ while keeping their resolutions as a preparation step. Then we fuse the features in three different ways at three different resolutions parallelly. As illustrated in Fig. \ref{fig.3}(a), when fusing at the resolution of $C_2$, features at smaller resolution will be upsampled to match the resolution of $C_2$, followed by a $1\times1~conv$. Similarlly in Fig. \ref{fig.3}(b) and (c), we adjust the features by either downsampling or upsampling to achieve the target resolution, followed by a $1\times1~conv$. After the resolutions are unified among the parallel features, we use an element-wise addition to fuse them at different resolutions parallelly. The fused features are then added back to the original pyramid features after a $1\times1~conv$. This parallel fusion structure gives us a chance to borrow information from other resolutions (either higher or lower) to output enhanced representations at every given resolution independently.

\textbf{Progressive Fusion.} After the parallel fusion, we further fuse the features of different resolution progressively. Starting from $C_5$, we upsample the features by a scale of 2, and then add them with the parallelly fused features at $C_4$ after a $1\times1~conv$. After the fusion is done at $C_4$ resolution, we take the similar steps to fuse at $C_3$ resolution, and then at $C_2$ resolutions. The fused features at different resolutions are all used in the next refinement stage. This progressive fusion structure allows us to enhance the features gradually and in a bottom-to-top way, which tries to maximize the capacity of the features at higher resolution to absorb more semantic comprehension from a lower resolution, especially for features at the highest resolution. 

\subsection{Feature Refinement}
We apply refinement to the fused features by using two modules: dilated bottleneck and attention module. This refinement is applied to features at all four resolutions. Then the features at $C_3$, $C_4$ and $C_5$ resolutions are upsampled to $C_2$ resolution, and concatenated together as the final input to a dilated bottleneck to predict the final heatmap.

\begin{figure}[!t]
    \centering
    \includegraphics[width=3in]{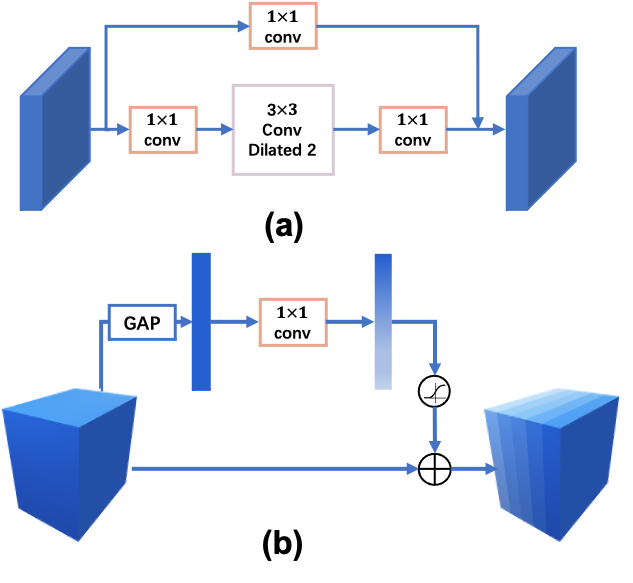}
    \caption{(a) Dilated Bottleneck with DilatedConv. (b) Partial Attention Module.}
    \label{fig.4}
\end{figure}

\textbf{Dialated Bottleneck} The structure of our proposed dilated bottleneck is shown in Fig. \ref{fig.4}(a). We use dilated convolution to enable us using smaller kernels of convolution to achieve a good trade-off between receptive field and complexity. 

\textbf{Attention Module} Attention mechanism has been proved successful in many applications. Here we use an attention module to weight the importance of different channels. As shown in Fig.~\ref{fig.4}(b), the module has two branches. The first branch uses Global Average Pooling, an 1x1 Conv and Sigmoid function to generate a weight vector. The second branch keeps the original input and multiple the weight from the first branch to output the attention-tuned feature maps.

\subsection{Loss Function}
During the process on training, the outputs of \textbf{Feature Extraction Stage} adopt $L2~Loss$ of all keypoints, at the same time the outputs of \textbf{Refinement Stage} adopt $L2^*~Loss$, which we select hard keypoints online based on traing loss(pixel-level heatmap $L2~Loss$). We only keep the top $\alpha$($\alpha$=10) keypoints with the largest loss among all \textit{N} keypoints.

\begin{figure*}[htb]
    \centering
    \includegraphics[scale=0.22]{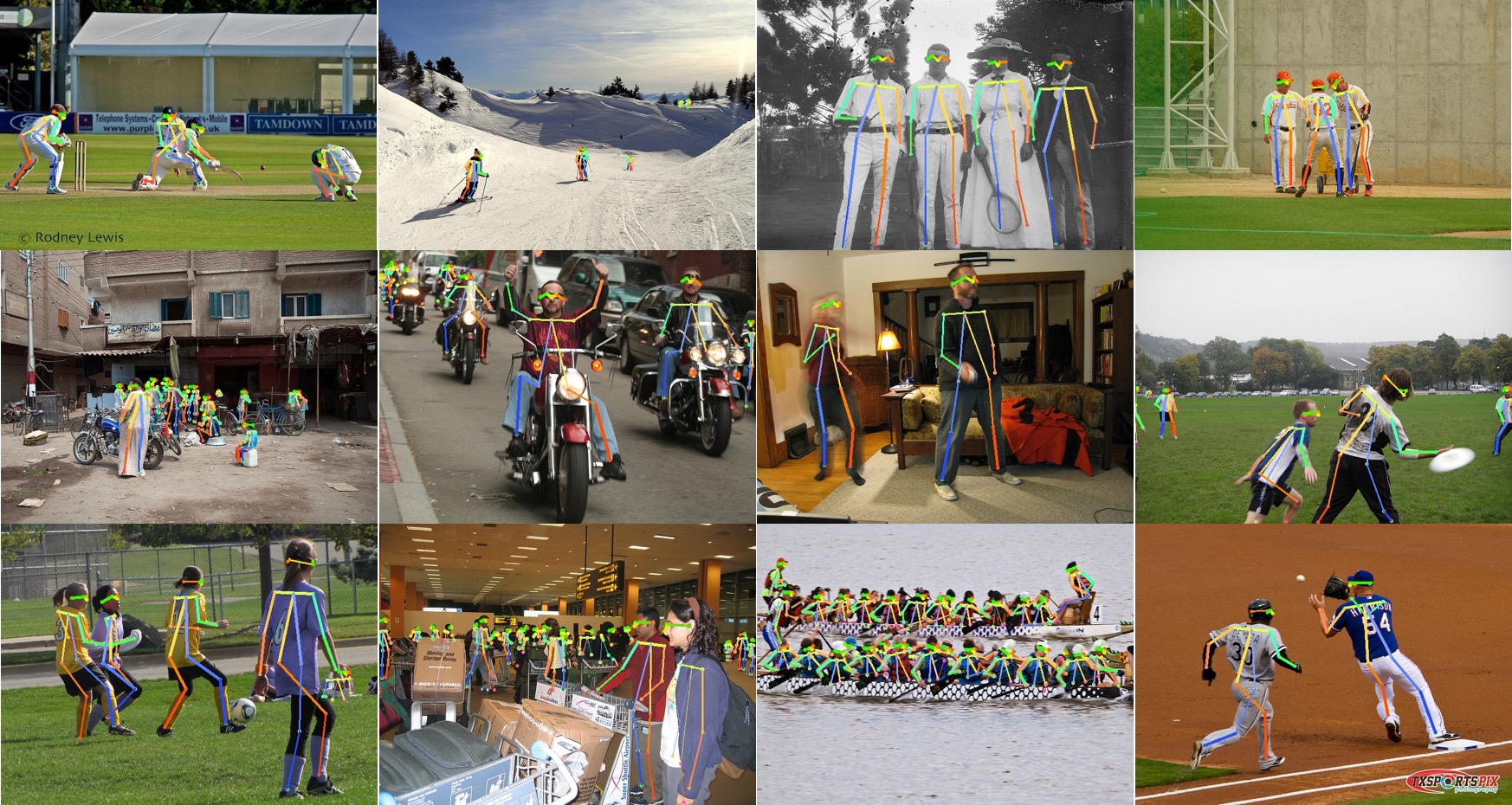}
    \caption{The visualization results on MSCOCO.}
    \label{fig.5}
\end{figure*}
\section{Experiments and Analysis}
We evaluate our proposed $P^2$ Net on two pose-estimation datasets. The overall results demonstrate that our framework achieves state-of-the-art verification accuracy across multi-person tasks.
\subsection{COCO keypoint Detection}
\subsubsection{Dataset}
Our models are trained on the MSCOCO \cite{lin2014microsoft} train dataset (includes 57K images and 150K person instances with 17 keypoints) and validated on MSCOCO val dataset (includes 5k images). The testing sets include test-dev sets (20K images)
\subsubsection{Training Details}
For each bounding-box, we crop the box from the image, which is resized to a fixed size, 384$\times$288. Then we adopt data augmentation policies previously searched by us. In order to get the auto data augmentation policies, we train from scratch with a global batch size 32, 384$\times$288 of images size, learning rate of 0.08, weight decay of $1^{e-4}$, $\alpha$ = 0.25 and $\gamma$ = 1.5 for the focal loss. 

We train 100 epochs, using stepwised decay where the learning rate is reduced by a factor of 10 at epochs 70 and 90. All models are trained on 8 GPUs over 3 days. Meanwhile, in order to save computing resources and speed up, we only randomly use 5k MSCOCO train images when searching for data augmentation sequences. The reward signal for the controller is the $mAP$ on MSCOCO val sets of 5,000 images. 

The models of pose estimation are trained using \textit{Adam} algorithm with an initial learning rate of 5$e^{-4}$. Note that we also decrease the learning rate by a factor of 2 every $3.6\times10^{6}$ iterations. We use a weight decay of 1$e^{-5}$ and set the training batch size 32. Batch normalization is used in our network. Generally, the training of ResNet-101 models takes about 2 days on 8 GPUs. Our models are initialized with weights of the ImageNet-pretrained model.

\subsubsection{Testing Details}
We apply a gaussian filter on the predicted
heatmaps, computing the heatmap by averaging the heatmaps of the original and flipped images. Each keypoint position is predicted by adjusting the highest calorific value position, which is shifted by a quarter in the direction from the highest response to the second highest response. We consider the product of boxes’ score and the average score of all keypoints as the final pose score of a person instance.
\begin{table*}[!t]
\footnotesize
\caption{Comparisons on COCO \textit{test-dev}.\#Params and FLOPs are calculated for the pose estimation network.}
\renewcommand{\arraystretch}{1.5}
\centering
    \setlength{\tabcolsep}{2mm}{
    \scriptsize
    \begin{tabular}{c|c|c|c|c|c|ccccc}
    \hline
    Method&Backbone&Input size&\#Params&GFLOPs&AP&$AP^{50}$&$AP^{75}$&$AP^{M}$&$AP^{L}$&AR\\
    \hline
    Mask-RCNN \cite{he2017mask}&ResNet-50-FPN&-&-&-&63.1&87.3&68.7&57.8&71.4&-\\
    G-RMI \cite{papandreou2017towards}&ResNet-101& 353$\times$257&42.6M& 57.0 &64.9&85.5& 71.3& 62.3 &70.0&69.7\\
    CPN \cite{chen2018cascaded}&ResNet-Inception& 384$\times$288& - &  - &72.1&91.4 &  80.0 & 68.7&77.2&78.5\\
    RMPE \cite{fang2017rmpe}&PyraNet& 320$\times$256& 28.1M&  26.7 &72.3&89.2 &  79.1 & 68.0&78.6&-\\
    CFN \cite{huang2017coarse}&-& -& -&-&72.6&86.1 &  69.7 & 78.3&64.1&-\\ 
    CPN \cite{chen2018cascaded} (ensemble)&ResNet-Inception& 384$\times$288& - &  - &73.0&91.7 &  80.9 & 69.5&78.1&79.0\\
    SimpleBaseline \cite{xiao2018simple}&ResNet-152& 384$\times$288& 68.6M &  35.6  &73.7&91.9&  81.1 & 70.3&80.0&79.0\\
    HRNet-W32 \cite{sun2019deep} & HRNet-W32 & 384$\times$288& 28.5M &  16.0  &74.9& 92.5 &  82.8 & 71.3 & 80.9 &80.1\\
    HRNet-W48 \cite{sun2019deep} & HRNet-W48 & 384$\times$288& 63.6M &  32.9  &75.5& 92.5 &  83.4 & 71.9 & 81.5 &80.5\\
    \textbf{Ours}&\textbf{ResNet101}&\textbf{384}\textbf{$\times$}\textbf{288}& \textbf{42.5M} &\textbf{26.3}&\textbf{77.3}&\textbf{93.1}&  \textbf{84.7}&\textbf{73.6}&\textbf{83.4}&\textbf{82.5}\\
    \hline
    \end{tabular}
    \label{tab.1}
    }
\end{table*}
\begin{table*}[!t]
    \caption{Performance comparisons on the MPII test set (PCKh@0.5).}
    \renewcommand{\arraystretch}{1.5}
    \centering
    \begin{tabular}{l|ccccccc|c}
    \hline
    Method & Hea&Sho&Elb&Wri&Hip&Kne&Ank&Total\\
    \hline
   Stack Hourglass \cite{newell2016stacked}. &98.2 &96.3& 91.2&87.1&90.1&87.4&83.6&90.9\\
   Sun et al \cite{sun2017human}. &98.1 &96.2& 91.2&87.2&89.8&87.4&84.1&91.0\\
   Chu et al \cite{chu2017multi}. &98.5 &96.3& 91.9&88.1&90.6&88.0&85.0&91.5\\
   Chou et al \cite{chou2018self}. &98.2 &96.8& 92.2&88.0&91.3&89.1&84.9&91.8\\
   Yang et al \cite{yang2017learning}. &98.5 &96.7& 92.5&88.7&91.1&88.6&86.0&92.0\\
   Ke et al \cite{ke2018multi}. &98.5 &96.8& 92.7&88.4&90.6&89.3&86.3&92.1\\
   Tang et al \cite{tang2018deeply}. &98.4 &96.9& 92.6&88.7&91.8&89.4&86.2&92.3\\
   SimpleBaseline \cite{xiao2018simple}&98.8 &96.6& 91.9&87.6&91.1&88.1&84.1&91.5\\
   HRNet-W32 \cite{sun2019deep}&98.6 &96.9& 92.8&89.0&91.5&89.0&85.7&92.3\\
   \textbf{Ours}&\textbf{98.8} &\textbf{97.0}& \textbf{93.9}&\textbf{89.9}&\textbf{92.1}&\textbf{92.0}&\textbf{86.4}&\textbf{92.9}\\
    \hline
    \end{tabular}
    \label{tab.2}
\end{table*}
\subsubsection{Experiment Results}
Tables~\ref{tab.1},\ref{tab.2} show the pose-estimation performance of our method. We get state-of-the-art results on both two datasets, MSCOCO test-dev and MPII. Our performance is better than HRNet, meanwhile the amount of parameters is much less than HRNet. Figure~\ref{fig.5} shows the visualization results of our method on the MSCOCO. We can see that we have achieved state-of-the-art visualization results on different scales of human body and different color pictures.
\subsection{Ablation Study} 
We analyze the effectiveness of each component in $P^2$ Net. Our net is evaluated on ResNet101 backbone.
\subsubsection{Detector Network}
Table~\ref{tab.9} shows the relationships between $mAP$ of detection and the corresponding $AP^*$ of the predicted keypoints, we have chosen the three detector networks, Faster R-CNN \cite{ren2015faster}, DetNet \cite{li2018detnet} and TridentNet \cite{li2019scale}. From the table, when the detection $AP$ increases and the accuracy of pose estimation has not been greatly improved. Therefore, the more important task for pose estimation is to enhance the accuracy of the keypoints rather than involve more boxes.
    \begin{table}[htb]
    \caption{ Comparing with different detector networks. $AP^*$ means the performance of $P^2$ Net based on the specific detector network. }
    \renewcommand{\arraystretch}{1.2}
    \centering
        \setlength{\tabcolsep}{0.3mm}{
        \begin{tabular}{c|c|c|c|c|c|c|c}
        \hline
        Detector&Backbone&mAP&$AP^{50}$&$AP^{S}$&$
        AP^{M}$&$AP^{L}$&$AP^*$\\
        \hline
        Faster R-CNN&ResNet50&37.4&59.0&18.3&41.7&52.9&\textbf{77.2}\\
        DetNet59&ResNet50&40.2&61.7&\textbf{23.9}&43.2&52.0&\textbf{77.3}\\
        TridentNet&ResNet101&\textbf{40.6}&\textbf{61.8}&23.0&\textbf{45.5}&\textbf{55.9}&\textbf{77.3}\\
        \hline
        \end{tabular}
        \label{tab.9}
        }
    \end{table}
\subsubsection{Differentiable Auto Data Augmentation ($DA^2)$}
As shown in Table~\ref{tab.6}, comparing our approach to the method based on our network without searching, we can see that performance has improved a lot. This indicates that auto data augmentation can effectively improve the performance and reduce noise induced by strategies designed by hand-crafted. $Ours^*$ means the method without searching, adopting the traditional data augmentations. 
    \begin{table}[htb]
    \footnotesize
    \caption{Comparing with the result without searching.}
    \renewcommand{\arraystretch}{1}
    \centering
        \setlength{\tabcolsep}{0.5mm}{
        \begin{tabular}{c|c|c|c|c|c|c|c}
        \hline
        Method&Backbone&AP&$AP^{50}$&$AP^{75}$&$
        AP^{M}$&$AP^{L}$&AR\\
        \hline
        \textbf{Ours}&\textbf{ResNet101}&\textbf{77.3}&\textbf{93.1}&  \textbf{84.7}&\textbf{73.6}&\textbf{83.4}&\textbf{82.5}\\
        $Ours^*$&ResNet101&75.8&91.7&83.1&72.2&81.5&82.2\\
        \hline
        \end{tabular}
        \label{tab.6}
        }
    \end{table}
\subsubsection{Attention Module}
As shown in Table~\ref{tab.7}, comparing our approach to the method based on our network without $AM$, we can see that performance has improved a lot. This indicates that $AM$ can gather features more selectivity and efficiently from shallow layers and deep layers. $Ours^*$ means the method without Attention Module, adopting the auto data augmentations. 
\begin{table}[htb]
    \footnotesize
    \caption{Compared with the result without Attention Module.}
    \renewcommand{\arraystretch}{1}
    \centering
        \setlength{\tabcolsep}{0.5mm}{
        \begin{tabular}{c|c|c|c|c|c|c|c}
        \hline
        Method&Backbone&AP&$AP^{50}$&$AP^{75}$&$
        AP^{M}$&$AP^{L}$&AR\\
        \hline
        \textbf{Ours}&\textbf{ResNet101}&\textbf{77.3}&\textbf{93.1}&  \textbf{84.7}&\textbf{73.6}&\textbf{83.4}&\textbf{82.5}\\
        $Ours^*$&ResNet101&76.2&92.5&83.1&72.2&82.2&81.4\\
        \hline
        \end{tabular}
        \label{tab.7}
        }
    \end{table}
    \begin{table}[htb]
    \footnotesize
    \caption{Compared with the result without parallel structures.}
    \renewcommand{\arraystretch}{1}
    \centering
        \setlength{\tabcolsep}{0.5mm}{
        \begin{tabular}{c|c|c|c|c|c|c|c}
        \hline
        Method&Backbone&AP&$AP^{50}$&$AP^{75}$&$
        AP^{M}$&$AP^{L}$&AR\\
        \hline
        \textbf{Ours}&\textbf{ResNet101}&\textbf{77.3}&\textbf{93.1}&  \textbf{84.7}&\textbf{73.6}&\textbf{83.4}&\textbf{82.5}\\
        $Ours^*$&ResNet101&76.3&92.6&82.9&70.4&81.6&83.6\\
        \hline
        \end{tabular}
        \label{tab.8}
        }
    \end{table}
\subsubsection{Parallel Fusion}
As shown in Table~\ref{tab.8}, comparing our approach to the method based on our network without the Parallel Fusion, we can see that the performance has improved a lot. This indicates that although the pyramid structure effectively balances the deep semantic information and the shallow spatial resolution, there still exists information loss. Meanwhile, we can also know that Parallel Fusion can make up for losses. $Ours^*$ means the method without Parallel Fusion, adopting the auto data augmentations. 
\section{Conclusion}
This paper proposes a top-down pose estimation framework with Differentiable Auto Data Augmentation. Our Differentiable Auto Data Augmentation ($DA^2$) method relaxes the discrete policy selection process to be differentiable using Gumbel-Softmax. To achieve efficient and accurate optimization, we propose a one-pass optimization strategy. A distinguishable method is to search for different extensions and strategies, which are regarded as different routines to be selected. Finally, Progressive  Fusion and Parallel Fusion aim to fuse feature maps of different levels of Pyramid Features so that high-level and low-level information can be utilized. Our proposed framework focuses on all stages of pose estimation, so it can greatly reduce the estimation error. Experimental results show that our proposed framework is superior to the state-of-the-art human pose estimation methods.
\section{Acknowledgements}
This work is partially funded by Beijing Natural Science Foundation (Grant No. JQ18017), Youth Innovation Promotion Association CAS (Grant No. Y201929) and DiDi GAIA Research Collaboration Initiative.

\bibliographystyle{IEEEtran}
\bibliography{egbib}
\end{document}